\documentclass[10pt,twocolumn,letterpaper]{article}
\usepackage{times}
\usepackage{epsfig}
\usepackage{multirow}
\usepackage{bbm}
\usepackage[normalem]{ulem}
\usepackage{subcaption}
\usepackage{xcolor}
\usepackage{color, colortbl}
\usepackage{caption}

\newcommand\methodshort{FOR }

\newcommand*{\boldone}{\text{\usefont{U}{bbold}{m}{n}1}}
\usepackage{siunitx}

\usepackage{wacv}              

\usepackage{graphicx}
\usepackage{amsmath}
\usepackage{amssymb}
\usepackage{booktabs}
\usepackage{wrapfig}

\usepackage[pagebackref,breaklinks,colorlinks]{hyperref}

\usepackage[capitalize]{cleveref}
\crefname{section}{Sec.}{Secs.}
\Crefname{section}{Section}{Sections}
\Crefname{table}{Table}{Tables}
\crefname{table}{Tab.}{Tabs.}

\def\wacvPaperID{1209} 
\def\confName{WACV}
\def\confYear{2025}

\begin{document}

\title{FOR: Finetuning for Object Level Open Vocabulary Image Retrieval}

\author{Hila Levi*\\
General Motors, RND, Israel\\
{\tt\small hila.levi@gm.com}
\and
Guy Heller*\\
General Motors, RND, Israel\\
{\tt\small guy.heller@gm.com}
\and
Dan Levi\\
General Motors, RND, Israel\\
{\tt\small dan.levi@gm.com}
}
\maketitle

\begin{abstract}
   As working with large datasets becomes standard, the task of accurately retrieving images containing objects of interest by an open set textual query gains practical importance. The current leading approach utilizes a pre-trained CLIP model without any adaptation to the target domain, balancing accuracy and efficiency through additional post-processing. In this work, we propose FOR: Finetuning for Object-centric Open-vocabulary Image Retrieval, which allows finetuning on a target dataset using closed-set labels while keeping the visual-language association crucial for open vocabulary retrieval. \methodshort is based on two design elements: a specialized decoder variant of the CLIP head customized for the intended task, and its coupling within a multi-objective training framework. Together, these design choices result in a significant increase in accuracy, showcasing improvements of up to 8 mAP@50 points over SoTA across three datasets. Additionally, we demonstrate that FOR is also effective in a semi-supervised setting, achieving impressive results even when only a small portion of the dataset is labeled.
\end{abstract}

\newcommand\blfootnote[1]{%
  \begingroup
  \renewcommand\thefootnote{}\footnote{#1}%
  \addtocounter{footnote}{-1}%
  \endgroup
}
\section{Introduction}\label{sec:intro}
\begin{figure}[t]
\centering
\includegraphics[width=1\linewidth]{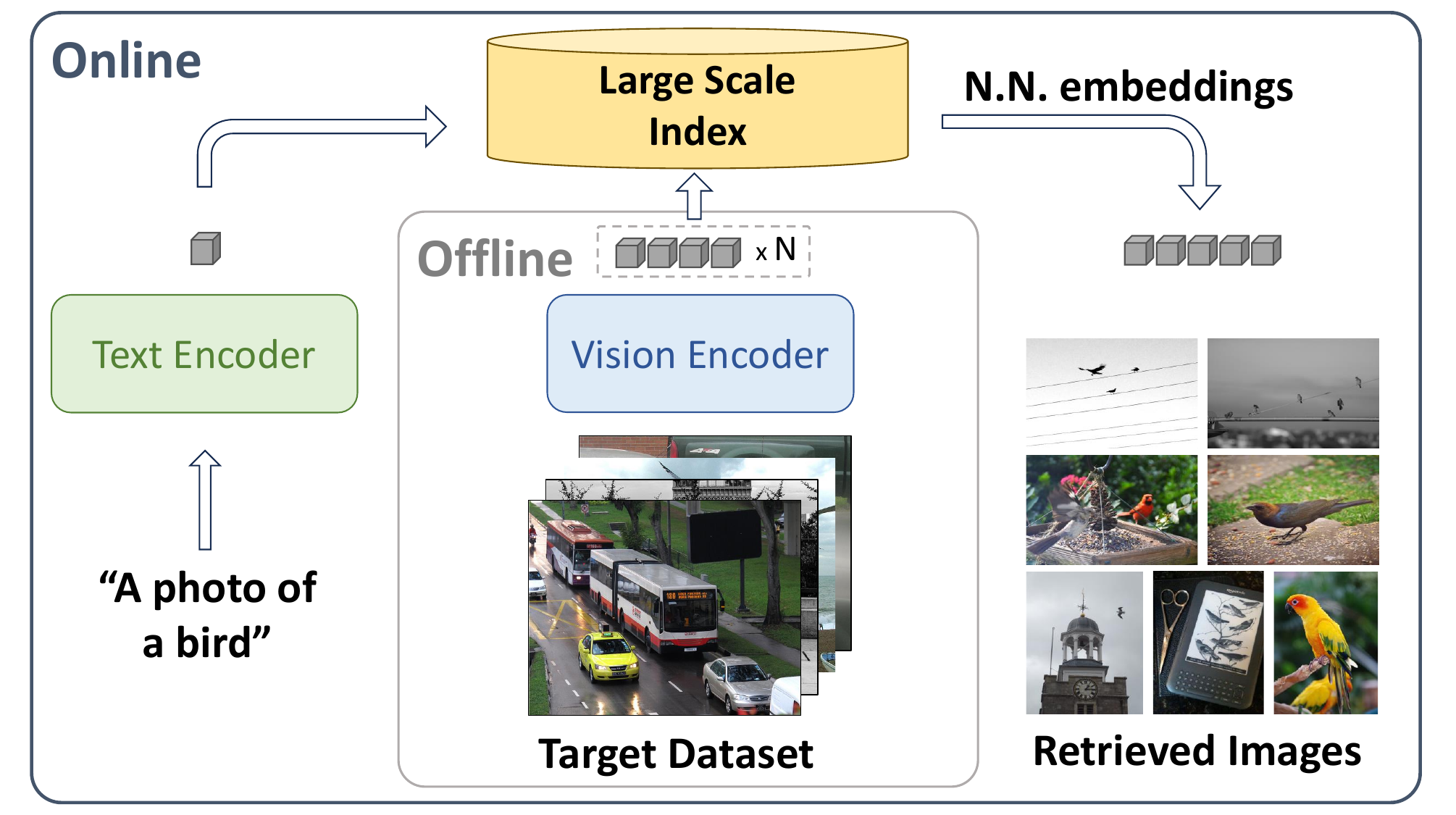}
\vspace{-0.5cm}
\caption{\footnotesize\textbf{Retrieval framework:} Images are first encoded with a predefined number of embeddings and stored in a large-scale index. Subsequently, rapid and repeatable retrieval is  performed by encoding text queries and conducting nearest neighbor searches. Notably, dual-encoder architectures enable separation into offline and online schemes, while scalability is enhanced by using a small number of embeddings per image.
} 
\vspace{-0.5cm}
\label{fig:retrieval_framework}
\end{figure}

Efficiently retrieving images containing objects of interest through on-demand open-set text queries is an important task in computer vision with diverse practical implications. Performing such targeted searches, especially over unlabeled rare concepts, facilitates tasks such as system performance evaluation, anomaly detection, and targeted data annotation. This capability is highly valuable in real-world applications, where dataset annotations are often incomplete or limited in scope due to the high costs of manual labeling. In such contexts, search methods excelling in a specific dataset hold significant practical value. 
\blfootnote{\null\hspace{-1.9em}$^\ast$ Equal contribution.}

While a number of early open-set retrieval algorithms are available (e.g., \cite{VSRN, chun2021pcme}), the advancement of open-vocabulary image retrieval has been significantly enhanced by the evolution of CLIP \cite{radford2021learning} and similar contrastive-based 
models (e.g.,  Florence \cite{yuan2021florence}, CoCa \cite{yu2022coca}). Trained on web-scale image-caption data, these frameworks generate a common embedding space for global image and caption representations. 
Their straightforward dual-encoder structure, with distinct vision and text encoders, facilitates retrieval through ranking the text-image similarity 
in a common embedding space and can be further scaled and accelerated\footnote{
Further acceleration can be achieved by using industrial search engines such as FAISS \cite{johnson2019billion}, which enable conducting nearest neighbor searches in a matter of milliseconds.}
by using frameworks as schematically illustrated in Figure \ref{fig:retrieval_framework}. Despite these advancements, CLIP's reliance on a single visual embedding (effective for image-caption matching), is insufficient for representing all objects in the image \cite{LeviHLF23}, as required in the object-centric open-vocabulary image retrieval (OC-OVIR) task. 

In a subsequent research trajectory, Dense-CLIP \cite{rao2022denseclip} was developed through a modification to CLIP's final pooling attention layer. It introduced a variant head to CLIP, facilitating the generation of dense embeddings while retaining the initial vision-language associations and has been employed across several applications (e.g., detection \cite{li2022adapting}, segmentation \cite{zhou2021denseclip}). Following, Dense-CLIP was adapted for object-based retrieval tasks in Cluster-CLIP \cite{LeviHLF23}, wherein a reduction in the number of output embeddings was achieved through the integration of supplementary CPU-based clustering algorithms.
This adaptation allows its usage within large-scale retrieval frameworks and significantly improves retrieval rates compared to CLIP, but offers no finetunning capability on a target dataset. The potential of such finetuning is well established in the context of the related problem of open vocabulary object detection (e.g., \cite{ViLD, owlvit}). However, as shown in \cite{LeviHLF23},
directly applying object detection methods to OC-OVIR is not scalable; each new query requires running the detection model on the entire dataset, thus demanding substantial computational resources. Alternatively, precomputing and storing all the visual embeddings requires significantly more storage.

In this work, we propose FOR, a framework for fine-tuning the image encoder of an open vocabulary model in order to enhance its OC-OVIR performance on a target dataset. The main challenges are twofold: first, effectively conducting fine-tuning using a limited set of labeled categories to enhance accuracy across all categories, including novel categories unseen during the training phase. Second, to create a representation that accurately captures image content with a limited number of embeddings, crucial for scalability and retrieval efficiency.

\begin{figure}[t]
\centering
\includegraphics[width=0.95\linewidth]{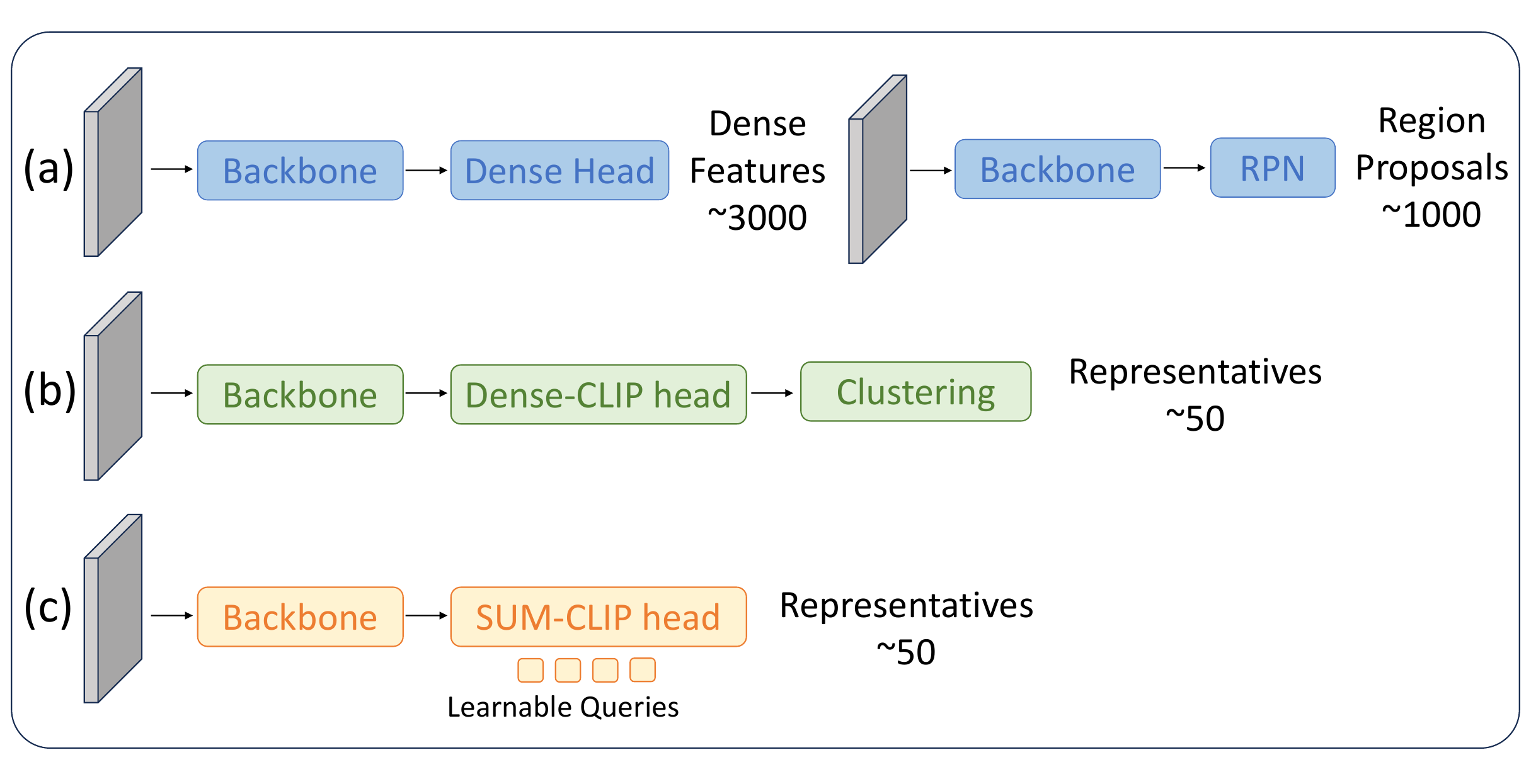}
\vspace{-0.2cm}
\caption{\footnotesize\footnotesize\textbf{Comparison of designs: } 
(a) Existing detection frameworks, either dense or RPN based, are impractical for retrieval due to their huge embedding representation; (b) Cluster-CLIP uses clustering for summarizing visual embeddings, but offers no finetunning capabilities;  (c) SUM-CLIP employs learnable queries and enables gradient flow. Consequently, SUM-CLIP achieves higher accuracy and faster inference times.}
\vspace{-0.5cm}
\label{fig:clip_vs_cluster}
\end{figure}

\begin{figure*}[t]
\centering
\footnotesize
\includegraphics[trim={1.5cm 4.2cm 1.5cm 4.2cm}, clip, width=0.95\linewidth]{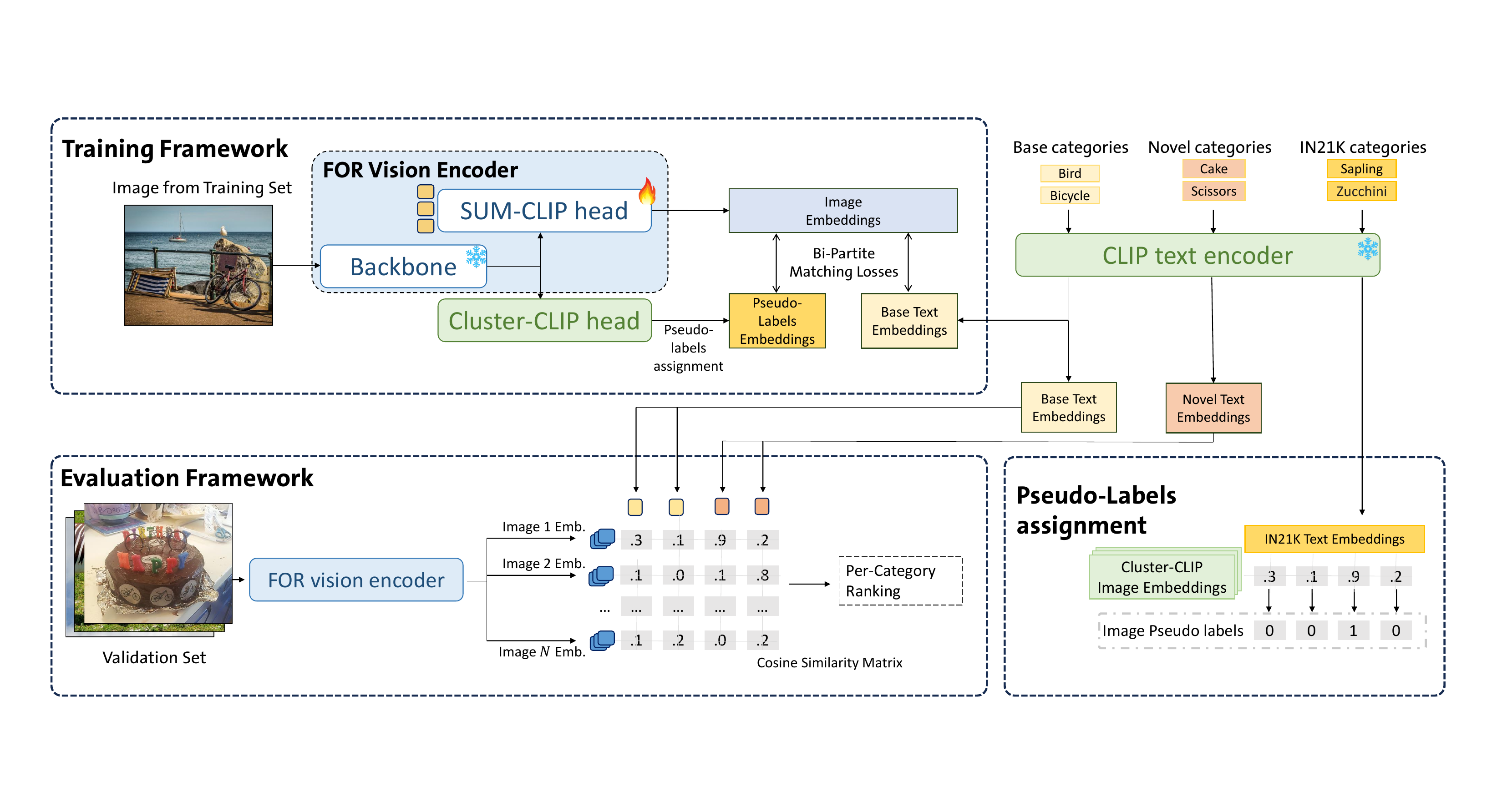}
\caption{\textbf{FOR overview.} Our training framework combines a supervised loss using the dataset base categories labels, and a pseudo-labels loss leveraging ImageNet-21K classes. Pseudo labels are assigned by filtering ImageNet-21K classes based on the similarity between their textual embeddings and the image embeddings from Cluster-CLIP. On inference, FOR can be used with any textual query, while base and novel labels are used solely for evaluation purposes.}
\vspace{-0.5cm}
\label{fig:method_framework}
\end{figure*}

To mitigate these challenges, our proposed framework employs a decoder variant CLIP head, termed SUM-CLIP (SUMmarizing image content with few embeddings), coupled within a multi-objective training scheme, depicted in Figure  \ref{fig:method_framework}. Drawing inspiration from Dense-CLIP, SUM-CLIP is also a variant of CLIP last attention layer, adapted for finetuning through additional learnable queries and decoder layers. Similar to Cluster-CLIP it generates a small set of representative embeddings per image while retaining CLIP's vision-language association via targeted freezing methodology. In contrast to Cluster-CLIP it allows finetuning, and eliminates the need of expensive CPU post-processing (as illustrated in Figure \ref{fig:clip_vs_cluster}).

Our multi-objective training scheme includes two branches. The first branch fine-tunes the SUM-CLIP head on a target dataset with closed-set vocabulary in a supervised manner. The second branch is used to compensate for open-vocabulary catastrophic forgetting. Specifically, we augment the conventional supervised finetuning with auxiliary targets in the form of pseudo-labels derived from Cluster-CLIP (current SoTA). Diverging from closed-set approaches, our pseudo-labels extend beyond the confines of dataset categories, allowing for more effective adaptation to unforeseen concepts. 

In the experiments, we show that employing our multi-objective framework significantly increases retrieval accuracy of novel categories by up to 8
mAP@50 points on three datasets while reducing visual inference time by a factor of three (mostly by eliminating the need for CPU post-processing and allowing batching) compared to Cluster-CLIP. Notably, pseudo-labels can extend to unlabeled data, typically found in larger quantities. We investigated leveraging unlabeled data by restricting supervised training to a small fraction of the labeled data, resulting in improved results compared to solely supervised methods. 

\vspace{0.3cm}
\noindent To summarize our contributions:

\begin{enumerate}
\item We introduce FOR, a first proposal of a fine-tuning framework for the OC-OVIR task, employing a multi-objective approach that enables learning without open-vocabulary forgetting.
\item We design SUM-CLIP, a CLIP-head decoder variant with learnable queries and targeted freezing methodology, tailored for the OC-OVIR task as it summarizes image content with few embeddings. 
\item We show the effectiveness of our approach by achieving significantly better results compared with current SoTA on three datasets: COCO \cite{LinMBHPRDZ14}, LVIS \cite{GuptaDG19}, and nuImages \cite{nuscenes2019}, increasing retrieval accuracy on novel categories by up to 8 mAP@50 points.
\end{enumerate}

\section{Related Literature}\label{sec:related}
Our work is closely related to cross-modal retrieval, open-vocabulary object-detection and semi-supervised learning. We briefly review related work in these domains.
\vspace{0.1cm}

\noindent\textbf{Cross-Modal Retrieval.} Aligning vision and language has a long-standing history of research \cite{misrm99:mori, frome2013devise, KirosSZ14, GongWHHL14, BottomUp}. The task is typically assessed using paired image-caption datasets, such as MS-COCO Caption \cite{LinMBHPRDZ14} and Flicker30K \cite{Flicker30K}. 
Over the years, a notable trend has been the expansion of training datasets; Early works \cite{VSRN, chun2021pcme, faghri2018vse++, KarpathyJL14, EngilbergeCPC18, 
SCAN, SAN, HuangWW17} 
relied on medium-sized datasets, while later efforts \cite{ALBEF, ViLBERT, PixelBERT, UNITER, OSCAR, LXMERT} emphasized Vision-Language (VL) pre-training on larger datasets, catalyzing the pre-training of VL models on massive web-scale corpora  (CLIP \cite{radford2021learning}, COCA \cite{yu2022coca}, and others \cite{LiT, yuan2021florence, jia2021scaling, SimVLM, GIT}), resulting in superior zero-shot performance across diverse datasets.  

Apart from the expansion of training datasets, literature can be categorized based on meta-architectures; One strand of research utilizes a joint image-text module (BLIP \cite{BLIP}, BLIP2 \cite{BLIP2} and others \cite{ALBEF, SAN, HuangWW17, UNITER, OSCAR}), which requires processing all images for each new text query, thereby restricting retrieval scalability. A more relevant approach employs a dual-stream architecture (e.g. VSRN \cite{VSRN}, PCME \cite{chun2021pcme} as early work exemplars, CLIP \cite{radford2021learning}, COCA \cite{yu2022coca} as recent examples), which can be integrated into the retrieval scheme illustrated in Figure \ref{fig:retrieval_framework}.  
In our work, we leverage recent advancements in cross-modal retrieval by using a customized CLIP-based architecture for object-centric image retrieval \cite{LeviHLF23, cai2022x, liu2022ovis}, capitalizing on CLIP's dual-encoder structure and extensive pretraining.

\noindent\textbf{Open-Vocabulary Object Detection.}
Recent studies in open-vocabulary object detection explore VL pre-training to detect objects beyond the base class vocabulary. However, solely finetuning the pretrained VL models often leads to open vocabulary forgetting. To address this challenge, one approach involves employing knowledge distillation to align region embeddings from a two-stages detector with CLIP VL features \cite{ViLD, regionCLIP, OADP, BARON, OC-OVD, F-VLM}. Furthermore, the effectiveness of utilizing region-text pseudo-labels, generated by leveraging datasets categories names \cite{DETIC, zhao2022exploiting, wu2023cora, OWLV2}, image captions \cite{regionCLIP, wu2023cora, OWLV2}, mosaics \cite{CLIM}, and phrase grounding \cite{regionCLIP}, has been demonstrated. For instance, CORA \cite{wu2023cora} generates pseudo bounding boxes using ImageNet-21K \cite{deng2009imagenet} categories or image caption data, while OWLV2 \cite{OWLV2} explores both curated label space from detection datasets and automatically generated N-grams from caption data on extensive unlabeled data.
Similarly, our framework employs knowledge distillation and pseudo-labels to mitigate open vocabulary forgetting in the OC-OVIR task. 
Notably, existing detection frameworks, including those
deemed dual encoders, are impractical for retrieval tasks due to their inefficient embedding representation; Two stages detector rely on large number ($\sim$ 1000) of region proposals to enhance novel category recall, while dense detectors utilize huge visual embedding space.

\noindent\textbf{Semi-supervised Learning.}
Learning from few labeled examples while effectively utilizing large amount of unlabeled data is a long-standing problem in machine learning \cite{zhu2005semi}. Current semi-supervised methods 
can be categorized into two main paradigms; The first incorporates unlabeled data as a regularization form during supervised learning \cite{sajjadi2016regularization, laine2016temporal, xie2020unsupervised}. 
The second utilizes pseudo labels \cite{lee2013pseudo}, generated through self-training \cite{arazo2020pseudo, pham2021meta, berthelot2019mixmatch, sohn2020fixmatch, xie2020self}, 
sometimes with additional pre-training \cite{zoph2020rethinking, chen2020big}, resulting in improved performance. Our work draws inspiration from the second paradigm, by leveraging Cluster-CLIP to generate pseudo-labels, which are then employed by our training framework, with or without unlabeled data. 
\section{Method}\label{sec:method}
\begin{figure*}[t]
\centering
\footnotesize
\includegraphics[width=0.95\linewidth]{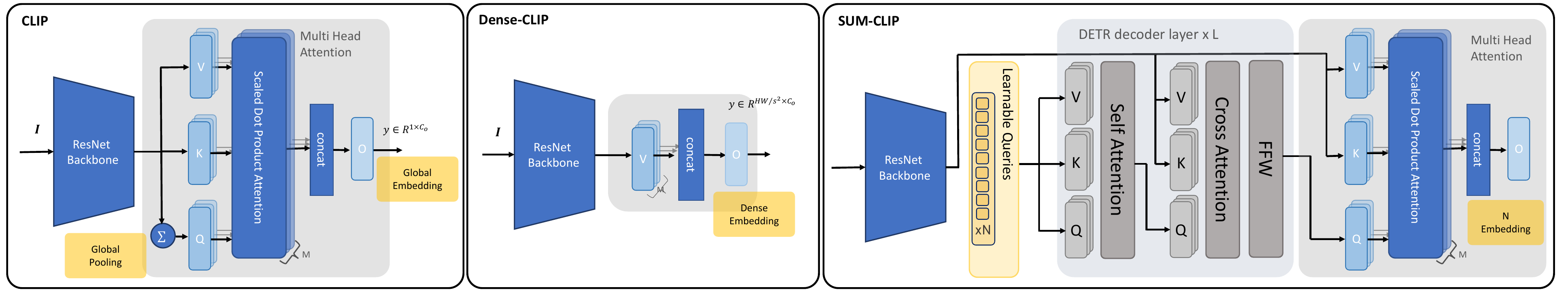}
\vspace{0.2cm}
\caption{\textbf{SUM-CLIP head}: CLIP (left) aims to represent the "average" semantics in images using $\bar{x}$ as a single query. Dense-CLIP (middle) focuses on local semantics induced by CLIP's original weights. SUM-CLIP (right) is designed to capture multiple objects by employing additional learnable queries and decoder layers preceding CLIP multi-head attention module.}
\vspace{-0.5cm}
\label{fig:method_summarization_head}
\end{figure*}

FOR aims to enable retrieval of images containing objects from novel categories beyond the base categories on which the visual encoder is finetuned. Formally, given a target dataset, FOR is trained on the target dataset training split with base-categories $C^B$ and evaluated on its evaluation split with both base categories $C^B$ and novel categories $C^N$ unseen through training ($C^B \cap C^N = \varnothing$). 

FOR architecture is built upon a pre-trained dual-encoder vision-language model (CLIP) with distinct processing pipelines for vision and text. Text processing involves applying the pre-trained CLIP text encoder to $C^B, C^N$, and additional pseudo-categories list $C^P$ defined in Section \ref{sec:met_pseudo}.
Visual processing (illustrated in Figure \ref{fig:method_framework}, left) utilizes a frozen CLIP ResNet backbone, followed by two parallel heads: a Cluster-CLIP head and a SUM-CLIP head (described in Section \ref{sec:method_heads}). Training adopts a semi-supervised approach, where the trainable SUM-CLIP head receives guidance from both supervised and pseudo-label losses, leveraging outputs from the Cluster-CLIP head (as detailed in Sections \ref{sec:met_pseudo}-\ref{sec:method_training_losses} and depicted in Figure \ref{fig:method_framework}, right).

During inference, the vision encoder, comprising of the CLIP backbone and the fine-tuned SUM-CLIP head, is applied to each image in the dataset, generating a set of embeddings per image that can be compared to any textual query. Evaluation is done by ranking the cosine similarity between the text embedding of each category in $C^B\cup C^N$ and the image embeddings. Similarly, interactive large-scale retrieval is facilitated by incorporating the finetuned vision encoder into a retrieval framework as depicted in Figure \ref{fig:retrieval_framework} and elaborated upon in Section \ref{sec:exp_qual}.

\subsection{SUM-CLIP Head}\label{sec:method_heads}
CLIP \cite{radford2021learning} vision processing pipeline includes a CLIP backbone followed by a multi-head pooling attention layer, designed to generate a single output embedding representing the "average semantics" of an image. 
SUM-CLIP builds upon recent studies (Dense-CLIP \cite{zhou2021denseclip} and Cluster-CLIP \cite{LeviHLF23}) by adapting the attention layer to produce image summarization with aggregated embeddings. The primary challenge lies in modifying the attention layer's structure to allow fine-tuning while preserving  the alignment between CLIP's vision and language components. Herein, we detail the implementations of CLIP, Dense-CLIP, and Cluster-CLIP, then introduce SUM-CLIP, a novel CLIP head variant that is tailored to the OC-OVIR task.

\vspace{0.3cm}
\noindent\textbf{CLIP:} The last layer in CLIP's visual encoder (Figure \ref{fig:method_summarization_head}, left) is implemented as a pooling multi-head attention layer, where the query is pooled from the input tensor through averaging. It sums information from all the pixels in the input tensor weighted by their similarity to the query and generates a single global embedding per image ($y \in R^{1 \times C_o}$):

\begin{align}\label{eq:clip}
& y = c \left( z \right),
& z = softmax \left( q(\bar{x}) \cdot k(X)^T \right) v(X) 
\end{align}

Here $X \in R^{K \times C_e}$ is the input tensor and $y \in R^{1 \times C_o}$ is the output embedding (one global vector of $C_o$ channels in the output representation of CLIP). 
$\bar{x}=\frac1K\sum_{i=1}^Kx_i$ represents the average of all spatial locations, $\{x_i\}_{i=1}^K$, of the input tensor $X$. $q: R^{C_e} \rightarrow R^{C_q}$, $k: R^{C_e} \rightarrow R^{C_q}$, $v: R^{C_e} \rightarrow R^{C_v}$ and $c: R^{C_v} \rightarrow R^{C_o}$ are respectively the query, key, value and output linear layers.

\vspace{0.3cm}
\noindent\textbf{Dense-CLIP:}
Dense-CLIP (Figure \ref{fig:method_summarization_head}, middle) produces dense patch embeddings aligned with CLIP's output space by  utilizing local semantics, already captured by the spatial locations at the input to CLIP's last attention layer. It is implemented by removing the query and key linear layers and substituting the value and output linear layers with 1x1 convolutional layers (initialized with CLIP weights) and formalized as:
\begin{align}\label{eq:denseclip}
& y_i = c \left( z_i \right),
& z_i = v(x_i)
\end{align}
here the output embedding $Y \in R^{K \times C_o}$ is a tensor, $y_i$ is the representation of its i’th spatial pixel: $Y=\{y_i \}_{i=1}^K$, $y_i \in R^{1 \times C_o}$. $K$, the number of output embeddings, is determined by the input image size and the model stride (e.g., $K=196$ for image size of $448 \times 448$ and stride 32).

\vspace{0.3cm}
\noindent\textbf{Cluster-CLIP:}
Suggested in \cite{LeviHLF23}, Cluster-CLIP aimed to improve Dense-CLIP scalability and adjust it for large-scale retrieval. It  produces aggregated embeddings by an additional aggregation module on top of Dense-CLIP embeddings. Formally, the aggregation module first clusters the dense features predicted by Dense-CLIP ($Y=\{y_i\}_{i=1}^K$) within $N$ clusters (e.g., by using K. Means), denoted as $\{C_j\}_{j=1}^N$, where $C_j \subset Y$ and $N<<K$. Then, it transfers one representative embedding per cluster (the average of the embeddings within the cluster) for future retrieval use. The main drawbacks of Cluster-CLIP are that it is unable to adapt to a specific dataset since it is non-trainable and that it requires computationally expensive post-processing. 

\vspace{0.3cm}
\noindent\textbf{SUM-CLIP (Ours):}
SUM-CLIP (Figure \ref{fig:method_summarization_head}, right) is designed to capture multiple objects ($Y \in R^{N \times C_o}$) by employing additional learnable
queries and decoder layers preceding CLIP last attention layer. The learnable queries, $Q \in R^{N \times C_e}$ ($N$ is pre-defined), are modulated, in an image dependent manner, by the decoder layers ($\tilde{Q} = F \left( Q, X \right)$), to produce the queries to the subsequent cross-attention layer, formulated as: 
\begin{align}\label{eq:sumclip}
& Y = c \left( Z \right),
Z = softmax \left( q(\tilde{Q}) \cdot k(X)^T \right) v(X) 
\end{align}
Where $X \in R^{K \times C_e}$ is the output featuremap of CLIP backbone ($K>>N$) and $q$, $k$, $v$ and $c$ are respectively the query, key, value and output linear layers of CLIP last attention layer. 
Notably, the additional decoder layers contribute to improved retrieval results, as evidenced in our experiments (\ref{sec:exp_ablt}). We hypothesize that aligning the queries with the image data, enabled by the additional decoder layers, is necessary for achieving favorable results.

With the additional learnable queries, two goals are met: first, the output dimension is limited to a small number of representatives, essential for large scale retrieval frameworks, without any additional post-processing. Second, as the linear layers of the last attention module can be initialized with CLIP weights, it allows training while mostly maintaining the original vision-language association of CLIP. 
Moreover, in the experiments (\cref{sec:exp_ablt}) we show a significant increase of zero shot retrieval accuracy while freezing the linear layers for several training setups.
We assume that, balancing between trainable capacity and pre-train knowledge, our task heavily favors stability over plasticity. 
Following, in all of our experiments, unless specified otherwise, the linear layers of the last attention module were kept frozen through training.

\subsection{Pseudo-label category assignment}\label{sec:met_pseudo}
Utilizing pre-trained neural networks to augment unlabeled data with pseudo-labels is a common strategy in semi-supervised research, proven to enhance accuracy across tasks and datasets. We source pseudo-label categories $C^P$ from ImageNet-21K \cite{deng2009imagenet} classes, which offers a diverse range of recognizable objects widely utilized in computer vision research. This approach extends beyond the limitations of close-set categories in the target dataset, ensuring a broad coverage that enhances the applicability of our methodology to real-world scenarios.

Given the categories set $C^P$, we assigned relevant pseudo-labels to each image through computing the softmax-normalized similarity between Cluster-CLIP branch output embeddings and the textual embeddings of $C^P$ (both $l_2$ normalized), followed by filtering categories with score lower than a predefined threshold (illustrated in \cref{fig:method_framework}, right). Specifically, given the visual embeddings from Cluster-CLIP output, $Y\in R^{N\times C_o}$, and the textual embeddings of $C^P$, $E \in R ^{|C^P|\times C_o}$, the probability of category $j$ appearing in an image, $p_j$, can be formulated as:
\begin{align}\label{eq:sumclip_2}
& S=E \cdot {Y}^T,
& p_j = \max_i\left(\{s_{ji}\}_{i=1}^N\right)
\end{align}
Where $S \in R^{|C^P|\times N}$ is a similarity matrix, $s_{ji}$ is the item in its $j$'th and $i$'th spatial location, and we omit $l_2$ and softmax normalization for clarity. Subsequently, pseudo-labels are selected if the probability is higher than a predefined threshold $p_j > th$. Several visual examples, along with their associated pseudo labels that demonstrate the benefits and limitations of the method are provided in the Supplementary Materials.

\subsection{Training losses}\label{sec:method_training_losses}

Our overall loss is a weighted sum of $\mathcal{L} = \gamma_{sup} \cdot \mathcal{L}_{sup} + \gamma_{pse} \cdot \mathcal{L}_{pse}$, 
where $\mathcal{L}_{sup}$ and $\mathcal{L}_{pse}$ are respectively the supervised loss and pseudo-label loss, calculated as set prediction loss against ground truth targets or pseudo labels. We adopt the set prediction loss as defined in DETR \cite{carion2020end} with necessary changes given the different tasks. Specifically\footnote{We define the set prediction loss for the supervised loss, with adjustments for the pseudo-label loss indicated in parentheses as necessary.}, given our predicted set of outputs $\hat{y}=\{\hat{y}_i\}_{i=1}^N$, $\hat{y}_i \in R^{1 \times C_o}$ and the ground truth set of categories $c=\{c_j\}_{j=1}^{T}$, $c_j \in C_B$ (or the pseudo-labels set of categories, $c_j \in C_P$), we use the Hungarian matching algorithm \cite{Kuhn1955Hungarian} to find the bipartite matching $\hat{\sigma}$ between the two sets $(c, \hat{y})$ which minimizes: 
\begin{equation}\label{eq:detr_match}
    \hat{\sigma} = \operatorname{argmin} \sum_{j=1}^{N} - \boldone_{c_j \neq \varnothing} \cdot \hat{p}_{\sigma (j)} (c_j)
\end{equation}
\noindent where, assuming $N$ is larger than $T$, $c$ is considered as a set of size N padded with $\varnothing$ (no object), and $\hat{p}_i(c_j)$ is the probability of class $c_j$ for $\hat{y}_i$ (the cosine similarity between $\hat{y}_i$ and the text embedding of class $c_j$ normalized by softmax across the classes). The loss is then defined as the cross entropy over matched items: 
\begin{equation}\label{eq:detr_loss}
    \mathcal{L} = \sum_{j=1}^{N} -w_{c_j} \log \hat{p}_{\hat{\sigma} (j)} (c_j) 
\end{equation}
\noindent where $w_{c_j}$ equals $1$ except when $c_j = \varnothing$, in which we set it to a small value to account for class imbalance (see details in the Supplementary Materials).
\section{Experiments}\label{sec:experiments}
In this section we verify the effectiveness of \methodshort for the task of OC-OVIR. Sections \ref{sec:exp_setup}-\ref{sec:exp_impl} details the datasets,  baselines, and implementation details of our methodology. Comparisons with existing methods and ablations are provided in Sections \ref{sec:exp_res}-\ref{sec:exp_ablt}. FOR's applicability in open-vocabulary semi-supervised settings is demonstrated in Section \ref{sec:exp_semi-sup}. Finally, Section \ref{sec:exp_qual} presents qualitative results of the complete interactive retrieval system.

\subsection{Experimental Setup}\label{sec:exp_setup}

\noindent\textbf{Datasets}.
Following previous conventions \cite{LeviHLF23}, we evaluate OC-OVIR on three publicly available datasets (COCO 2017 \cite{LinMBHPRDZ14}, LVIS \cite{GuptaDG19} and nuImages \cite{nuscenes2019}), using the datasets' semantic categories as queries. Each dataset's categories are divided into base categories, used for training and evaluation, and novel categories, reserved solely for evaluation. Specifically, for  COCO, a widely used dataset for object detection, we adhere to the practice used in COCO-OVD benchmark \cite{DBLP:conf/eccv/BansalSSCD18} and divide the dataset's categories into 48 base and 17 novel categories. LVIS is a benchmark dataset for long-tail object recognition, annotated with 1,203 semantic categories that are divided into frequent, common and rare. Following the conventions in \cite{regionCLIP, owlvit}, rare categories are treated as novel, while the rest are considered base categories. nuImages is a large-scale public dataset for autonomous driving. We define 8 rare categories (appearing in less than 5\% of the images) as novel and 10 frequent categories as base.

\noindent\textbf{Evaluation Protocol}. 
To evaluate our method we follow the evaluation protocol defined in \cite{LeviHLF23}. Specifically, the trained model is first used to create embeddings for each image. Then, images are sorted for each dataset's category based on their maximal similarity over all of their embeddings. We report $mAP@50$ (defined in \cite{kaggle2022gupta} and used in \cite{LeviHLF23, liu2022ovis}) which considers the top 50 images only. Images are considered \textit{true positive} for a category if they contain an object of that category.

\noindent\textbf{Baselines}. 
We compare against existing methods for OC-OVIR, namely CLIP, Dense-CLIP and Cluster-CLIP, the later being the current SoTA. Additionally, when possible, we compare to dual-stream caption-based retrieval methods, finetuned on COCO dataset with caption annotations. Specifically, we compare to PCME \cite{chun2021pcme} and VSRN \cite{VSRN}, as early methods predating CLIP, and to BLIP2 \cite{BLIP2} and CoCa \cite{yu2022coca}, as concurrent or subsequent to CLIP. To enable comparison, we eliminate the re-ranking stage of BLIP2\footnote{BLIP2  uses a joint image-text encoder and adapt it to retrieval by applying a dual encoder variant followed by reranking with the joint encoder.}.    

\subsection{Implementation Details}\label{sec:exp_impl}
We used the ResNet-50x64 CLIP backbone from the CLIP library \cite{DBLP:conf/icml/RadfordKHRGASAM21}. Models were initialized with CLIP weights, with the exception of SUM-CLIP's decoder layers and learnable queries, which were randomly initialized using the Xavier uniform distribution \cite{glorot2010understanding}. Unless specifically mentioned, SUM-CLIP uses 2 decoder layers (following the DETR decoder layers architecture \cite{carion2020end}) and 50 learnable queries with a dimension of 4096 per query. All experiments were conducted using a single Nvidia GPU. 

We trained FOR using pseudo-label and supervised losses with equal weights ($\gamma_{pse}=\gamma_{sup}=1$) for the COCO and LVIS datasets, and $\gamma_{pse}=10$ for nuImages. Training was conducted for 25 epochs using the Adam optimizer \cite{kingma2014adam}, a dropout ratio of 0.1, and an initial learning rate of $10^{-5}$, which decayed by a factor of 0.1 after 15 epochs. Cluster-CLIP was used with K-Means clustering, 50 clusters,  and default parameters suggested in \cite{LeviHLF23}. The confidence threshold for the pseudo labels was chosen through an hyper-parameter search to be \textit{$5e^{-4}$}. Further implementation details are provided in the Supplementary Materials. 

\begin{table*}[t]
\begin{center}
\resizebox{\textwidth}{!}{%
\begin{tabular}{lccc|cc|ccc|ccc|ccc}
\hline
& & & & 
\multicolumn{2}{c|}{losses} &
\multicolumn{3}{c|}{COCO - mAP@50} & \multicolumn{3}{c|}{LVIS - mAP@50} & \multicolumn{3}{c}{nuImages - mAP@50}\\
Method & FT. & $\#$rep. & CPU-PP & sup. & p.l. & base & novel & all & base & novel & all & base & novel & all \\
\hline\hline
\multicolumn{4}{l|}{\textbf{\textit{retrieval methods fine-tuned on coco-caption}}} & \multicolumn{2}{c|}{} & \multicolumn{3}{c|}{} & \multicolumn{3}{c|}{} & \multicolumn{3}{c}{} \\
VSRN, ResNet-101 \cite{VSRN}  & $\triangle$\footnotemark{} & 1 & $\times$ &
\multicolumn{2}{c|}{\textcolor{lightgray}{-}} &
69.39 & 76.26 & \textcolor{lightgray}{71.19} & 46.55 & 20.46 & \textcolor{lightgray}{42.09} & \textcolor{lightgray}{-} & \textcolor{lightgray}{-} & \textcolor{lightgray}{-} \\
PCME, ResNet-152 \cite{chun2021pcme}  & $\triangle$\textcolor{red}{\footnotemark[\value{footnote}]} & 7 & $\times$ & \multicolumn{2}{c|}{\textcolor{lightgray}{-}} &
67.89 & 74.87 & \textcolor{lightgray}{69.72} & 51.37 & 27.70 & \textcolor{lightgray}{47.32} & 
60.37 & 1.2 & \textcolor{lightgray}{34.07}\\

CoCa, ViT-L, \cite{yu2022coca} & $\triangle$\textcolor{red}{\footnotemark[\value{footnote}]} & 1 & $\times$ & 
\multicolumn{2}{c|}{\textcolor{lightgray}{-}} &
73.28 & 80.35 & \textcolor{lightgray}{75.13} & 65.41 & 47.85 & \textcolor{lightgray}{62.40} & 86.30  & 10.96 & \textcolor{lightgray}{52.82} \\

BLIP2, ViT-g \cite{BLIP2} & $\triangle$\textcolor{red}{\footnotemark[\value{footnote}]} & 32 & $\times$ & 
\multicolumn{2}{c|}{\textcolor{lightgray}{-}} &
78.04 & 84.50 & \textcolor{lightgray}{79.73} & 67.62 & 56.27 & \textcolor{lightgray}{65.68} & 95.71 & 13.40 & \textcolor{lightgray}{59.14} \\

\hline
\multicolumn{4}{l|}{\textbf{\textit{object-centric retrieval methods}}} & \multicolumn{2}{c|}{} & \multicolumn{3}{c|}{} & \multicolumn{3}{c|}{} & \multicolumn{3}{c}{} \\

CLIP, RN50x64 & $\times$ & 1 & $\times$ & 
\multicolumn{2}{c|}{\textcolor{lightgray}{-}} &
68.56 & 77.68 & \textcolor{lightgray}{70.95} & 64.54 & 53.03 & \textcolor{lightgray}{62.56} & 69.22 & 8.60 & \textcolor{lightgray}{42.29}\\

Dense-CLIP, RN50x64 & $\times$ & 196 & $\times$& 
\multicolumn{2}{c|}{\textcolor{lightgray}{-}} &
76.25 & 81.10 & \textcolor{lightgray}{77.52} & 73.53 &  57.90 & \textcolor{lightgray}{70.85} & 70.10 & \textcolor{blue}{14.47} & \textcolor{lightgray}{45.37}\\

Cluster-CLIP, K.M. RN50x64 & $\times$ & 25 & \checkmark & 
\multicolumn{2}{c|}{\textcolor{lightgray}{-}} &
 77.81 & 84.44 & \textcolor{lightgray}{79.55} & 70.34 & 54.55 & \textcolor{lightgray}{67.63} & 79.62 & 13.43 & \textcolor{lightgray}{50.20} \\

Cluster-CLIP, K.M. RN50x64 & $\times$ & 50 & \checkmark & 
\multicolumn{2}{c|}{\textcolor{lightgray}{-}} &
76.29 & 82.60 & \textcolor{lightgray}{77.94} & 71.79 & 56.55 & \textcolor{lightgray}{69.18} & 80.10 & 13.76 & \textcolor{lightgray}{50.63} \\
\hline

\multicolumn{4}{l|}{\textbf{\textit{ours}}} & \multicolumn{2}{c|}{} & \multicolumn{3}{c|}{} & \multicolumn{3}{c|}{} & \multicolumn{3}{c}{} \\

\methodshort, RN50x64 & \checkmark & 50 & $\times$ & 
\checkmark & $\times$ &
{91.11} & 21.51 & \textcolor{lightgray}{72.91} & {82.49} & {59.97} & \textcolor{lightgray}{78.64} & {93.39}	& 3.01 & \textcolor{lightgray}{45.63} \\

\methodshort, RN50x64 & \checkmark & 50 & $\times$ &  
$\times$ & \checkmark & 
81.69 & \textcolor{blue}{87.54} & \textcolor{lightgray}{83.22} & 75.05 & 58.85 & \textcolor{lightgray}{72.28} & 85.63 & 12.03 & {\textcolor{lightgray}{52.92}}\\ 

\arrayrulecolor{lightgray}\hline

\rowcolor[rgb]{ .949,  .949,  .949}\methodshort, RN50x64 & \checkmark & 25 & $\times$ & 
\checkmark & \checkmark & 
 88.97 & \textcolor{red}{89.17} & \textcolor{lightgray}{89.02} & 78.84 & \textcolor{blue}{60.14} & \textcolor{lightgray}{75.64} & {91.71} & 12.12 & \textcolor{lightgray}{56.34}\\

\rowcolor[rgb]{ .949,  .949,  .949}\methodshort, RN50x64 & \checkmark & 50 & $\times$ & 
\checkmark & \checkmark & 
88.14 & \textcolor{red}{89.17} & \textcolor{lightgray}{88.41} & {81.21} & \textcolor{red}{64.15} & \textcolor{lightgray}{78.29} & 91.28 & \textcolor{red}{15.09} & \textcolor{lightgray}{57.42}\\

\arrayrulecolor{black}\hline
\end{tabular}}
\end{center}
\vspace{-0.5cm}
\caption{Evaluation results on COCO2017, LVIS and nuImages val sets. First and second best scores are marked in \textcolor{red}{red} and \textcolor{blue}{blue}. \methodshort demonstrates high retrieval accuracy with low computational and memory cost. }
\label{tab:dense_cocolvis_wlosses}
\vspace{-0.5cm}
\end{table*}

\subsection{Results} \label{sec:exp_res}
Table \ref{tab:dense_cocolvis_wlosses} presents the retrieval performance of \methodshort compared to the baselines. The `FT' column denotes methods that are fine-tuned on the target dataset, the `\#rep' column indicates the number of embeddings per image, and the 
`CPU-PP' column marks methods that require significant CPU post processing. 
Evidently, \methodshort with 50 representatives (last line) surpasses Cluster-CLIP, the current SoTA, with a significant gain. It improves upon Cluster-CLIP by up to 7.6 mAP@50 points on novel categories and 10.3 mAP@50 on base categories, without expensive CPU post-processing. When reducing the number of queries to 25, \methodshort still outperforms Cluster-CLIP in two out of three benchmarks and achieves comparable results in the third.

Table \ref{tab:dense_cocolvis_wlosses} also presents results using only supervised loss or pseudo-label loss (fourth and third lines from the bottom). Notably, using only supervised loss leads to catastrophic forgetting of novel classes on COCO and nuImages, resulting in a 68 mAP@50 point reduction in novel categories on COCO. In contrast, on LVIS, supervised learning achieves high results for novel categories as well. We hypothesize that this difference is due to the increased volume and semantic connections among novel and base categories in LVIS, where images featuring uncommon objects might be identified based on their semantic correlation with base categories. Remarkably, while using only pseudo-labels falls short of dual-loss training, it still outperforms Cluster-CLIP in most benchmarks. Finally, using both losses (last row) improves results across all benchmarks.

FOR is based on two design elements: a specialized CLIP head variant (SUM-CLIP), and its coupling within a multi-objective training framework. To demonstrate the decoupled effectiveness of these elements, Table \ref{tab:finetuned} presents the results of finetuning CLIP and Dense-CLIP heads for the OC-OVIR task, employing an identical training approach as FOR. 
While fine-tuning Dense-CLIP improves its performance and showcases the effectiveness of the FOR framework, utilizing a SUM-CLIP head surpasses these results using only a quarter of the embeddings. Notably, fine-tuning CLIP within our training approach (following a full hyper-parameter search) decreased performance on novel categories. 
We hypothesize that CLIP, which generates a single output embedding, loses relevant pseudo-label training signals due to its inherent "one-to-one" correspondence. 
\begin{table}[h]
\captionsetup{font=footnotesize}
\begin{center}
\resizebox{0.95\linewidth}{!}{%
\footnotesize
\begin{tabular}{lcc|ll|ll}
\hline
& & & \multicolumn{2}{c|}{COCO} & \multicolumn{2}{c}{LVIS} \\
Method & fine-tuned & \#$rep$. & base & novel & base & novel \\
\hline\hline
CLIP & $\times$ & 1 & 68.56 & 77.68 & 64.54 & 53.03\\
FT-CLIP & \checkmark & 1 & 77.67 & 74.60 & 63.53 & 46.89 \\
\hline
Dense-CLIP & $\times$ & 196 & 76.25 & 81.10 &73.53 & 57.90\\
FT-Dense-CLIP & \checkmark & 196 &  88.00 & 86.67 & 77.46 & 61.61 \\
\hline
FOR, ours & \checkmark & 50 & 88.14 & \textbf{89.17} & 81.21 & \textbf{64.15}\\
\hline
\end{tabular}}
\end{center}
\vspace{-0.55cm}
\caption{Fine-tuning CLIP and Dense-CLIP.}
\label{tab:finetuned}
\vspace{-0.3cm}
\end{table}

\subsection{Ablations}\label{sec:exp_ablt}
In this section, we perform ablation studies to investigate the impact of different configurations in our proposed model. Experiments are evaluated on the COCO2017 val set, replicating the training procedure and hyperparameters used for the main results unless specified otherwise.
\footnotetext{Methods are fine-tuned on COCO dataset with caption annotations.}

\noindent\textbf{Pseudo Labels.}
Our pseudo-labeling strategy utilizes ImageNet-21K categories, as a widely accepted external knowledge source, with a possible partial overlap with the target dataset's novel categories. Table \ref{tab:pseudo} shows retrieval results for training our system with (third column checked $\checkmark$) or without  utilizing the overlapping novel categories for pseudo-labels creation. Our analysis indicates that leveraging this extensive pseudo-labeling strategy, whether including (row 4) or excluding (row 3) the 'novel' component, improves novel category retrieval by 6-7.5 points
\begin{table}[h]
\captionsetup{font=footnotesize}
\begin{center}
\resizebox{0.95\linewidth}{!}{
\begin{tabular}{l|ccc|ll|ll}
\hline
& \multicolumn{3}{c|}{losses} & \multicolumn{2}{c|}{COCO} & \multicolumn{2}{c}{LVIS} \\
Method & sup. & IN21/novel & novel & base & novel & base & novel \\
\hline\hline
Cluster-CLIP & $\times$ & $\times$ & $\times$ & 76.29 & 82.60 & 71.79 & 56.55\\
FOR, sup. only & \checkmark & $\times$ & $\times$ & 91.11 & 21.51 & 82.49 & 59.97\\
FOR, IN21/novel & \checkmark & \checkmark & $\times$ & 87.71 & 87.66 & 81.01 & 62.83\\
FOR, IN21K & \checkmark & \checkmark & \checkmark & 88.14 & \textbf{89.17} & 81.21 & {64.15}\\
\hline
\end{tabular}}
\end{center}
\vspace{-0.55cm}
\caption{Ablations on pseudo-labels}
\label{tab:pseudo}
\vspace{-0.3cm}
\end{table}
compared to Cluster-CLIP, validating the effectiveness of our approach.

\noindent\textbf{Freezing methodology.}
Table \ref{tab:ablations-freezing} compares different freezing methods for the SUM-CLIP head under various learning settings. Specifically, we compare learning with only supervised loss and no decoder layers (left column), learning with supervised and pseudo-label losses with no decoder layers (middle column), and learning with supervised and pseudo-label losses with two decoder layers (right column).
In all cases, unfreezing the value and output linear layers (top row) hinders the model's ability to learn effectively, indicating the challenge in maintaining CLIP's visual-textual association. When pseudo labels are used with no decoder layers, it is beneficial to increase network capacity by unfreezing the query and key linear layers (center, middle). Lastly, \methodshort with two decoder layers and frozen linear layers (right, bottom) achieves the highest overall score. 

\begin{table}[h]
\captionsetup{font=footnotesize}
\scriptsize
\centering
\begin{tabular}{c|cc|cc|cc}
\hline
& \multicolumn{2}{c|}{sup, 0 layers} & \multicolumn{2}{c|}{p.l., 0 layers} & \multicolumn{2}{c}{p.l., 2 layers} \\
 frozen & base & novel & base & novel & base & novel\\
\hline\hline
none & 42.56 & 5.35 & 61.21 & 74.97 & 61.95 & 69.55 \\
v,o & 83.59 & 17.91 & 86.43 & 87.31 & 88.65 & 87.93 \\
\hspace{0.25cm}  q,k,v,o \hspace{0.25cm}   & 75.10 & 80.12 & 78.19 & 83.96 & 88.14 & \textbf{89.17} \\
\hline
\end{tabular}
\caption{Ablation on the freezing methodology}
\label{tab:ablations-freezing}
\end{table}

\begin{table*}[]
\scriptsize
\begin{center}
\resizebox{0.85\textwidth}{!}{%
\begin{tabular}{l|cc|cc|cc|cc|cc|cc}
\hline
\multicolumn{1}{l|}{\multirow{2}{*}{Method}}& 
\multicolumn{2}{c|}{0.5\%} & 
\multicolumn{2}{c|}{1\%} &
\multicolumn{2}{c|}{2\%} & 
\multicolumn{2}{c|}{5\%} & 
\multicolumn{2}{c|}{10\%} & 
\multicolumn{2}{c}{100\%}\\
 & base & novel & base & novel & base & novel & base & novel & base & novel & base & novel \\
\hline\hline
Supervised  & 81.35 & 13.65 & 85.72 & 13.53 & 88.62 & 13.40 & 90.40 & 13.88 & 90.83 & 14.93 & 91.11 & 21.51 \\
FOR & 83.48 & 87.92 & 84.44 & 88.26 & 85.39 & 88.32 & 86.36 & 88.27 & 86.74 & 88.04 & 88.14 & 89.17 \\
\hline
\end{tabular}}
\end{center}
\vspace{-0.5cm}
\caption{Semi-Supervised evaluation results on COCO-2017.}
\label{tab:semi-sup}
\end{table*}

\noindent\textbf{Number of queries.}
Table \ref{tab:ablations-n-queries} illustrates the impact of query quantity on retrieval performance. Using 5-10 queries tends to specialize in base categories at the expense of open-vocabulary capabilities. Increasing the query count to 25 yields comparable results to those obtained with 50 queries. This aligns with prior research \cite{LeviHLF23} showing that  effective object-centric image retrieval on the COCO dataset can be achieved with a relatively small number of categories.

\noindent\textbf{Number of decoder layers.}
Table \ref{tab:ablations-n-layers} examines the impact of the number of decoder layers in the SUM-CLIP head. Notably, SUM-CLIP struggles in both base and zero-shot categories when no decoder layers are used. Adding a single decoder layer significantly improves retrieval results. We hypothesize that the gain is owed to the fact that additional image information is processed by the decoder layer, enabling the model to infer image specific queries. Lastly, two layers provide the best overall optimal performance, while further increasing network capacity decreases it.

\begin{table}[h]
\captionsetup{font=footnotesize}
\centering
\scriptsize
\begin{minipage}[t]{0.48\linewidth}
\centering
\begin{tabular}{ccc}
\hline
 $\#$queries & base & novel \\
\hline
 5  & 90.72 & 78.62 \\
 10 & 89.63 & 85.12 \\
 25 & 88.97 & \textbf{89.17} \\
 50 & 88.14  & \textbf{89.17}  \\
\hline
\end{tabular}
\caption{Number of queries}
\label{tab:ablations-n-queries}
\end{minipage}
\hfill
\begin{minipage}[t]{0.48\linewidth}
\centering
\begin{tabular}{ccc}
\hline
 $\#$layers & base & novel \\
\hline
 0 & 78.19 & 83.96  \\
 1 & 87.84 & 88.35 \\
 2 & 88.14 & \textbf{89.17} \\
 3 & 87.96 & 88.54 \\
 \hline
\end{tabular}
\caption{Number of decoder layers}
\label{tab:ablations-n-layers}
\end{minipage}
\end{table}

\vspace{-0.3cm}
\subsection{Open-Vocabulary Semi-Supervised Results} \label{sec:exp_semi-sup}
To the best of our knowledge, there is no established evaluation protocol for semi-supervised open-vocabulary retrieval. Thus, we adopt a common approach from closed-set semi-supervised object detection \cite{DBLP:journals/corr/abs-2005-04757}. Specifically, we randomly sample 0.5, 1, 2, 5, and 10\% of the train dataset as labeled data and use the remainder as unlabeled. In our implementation, unlabeled data was used only with the pseudo-labels loss $\mathcal{L}_{pse}$, while labeled data was used with both $\mathcal{L}_{sup}$ and $\mathcal{L}_{pse}$. For each labeling regime, we report the average results over 5 folds.

Table \ref{tab:semi-sup} presents results for \methodshort and a supervised-only baseline (using $\mathcal{L}_{sup}$ only) on the COCO dataset. Notably, \methodshort achieves comparable results to the fully labeled regime on novel categories with 1\%-10\% of the labeled data, showcasing the efficacy of the pseudo-labels loss. For base categories, FOR consistently improves results with the increase in labeled data. Additionally, \methodshort significantly surpasses the supervised baseline in novel categories, showing improvements of up to 75 mAP@50 points.

\subsection{System Level Qualitative Results}\label{sec:exp_qual}
\begin{figure}[h]
\centering
\includegraphics[trim={0 8.3cm 0 0},clip, width=0.8\linewidth]{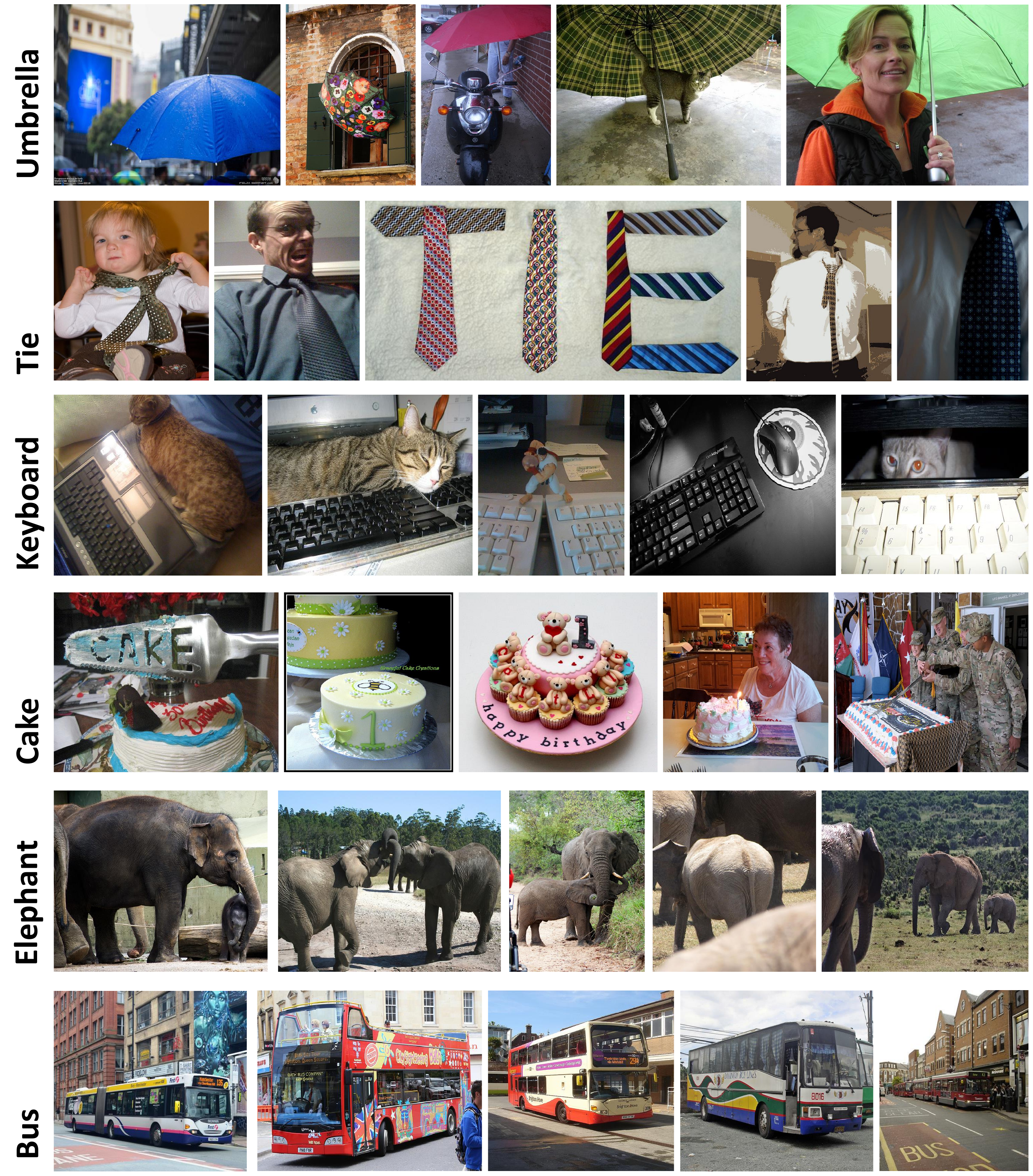}
\caption{\footnotesize\textbf{Qualitative Examples:} Top-5 retrieved images from our overall system using COCO novel classes as queries. The index was created from 40K unlabeled COCO images with SUM-CLIP.}
\vspace{-0.5cm}
\label{fig:qualitative_results}
\end{figure}

Figure \ref{fig:qualitative_results} illustrates qualitative results for selected COCO novel category names (unseen during training), acquired by utilizing a fine-tuned SUM-CLIP head on the unlabeled segment of the COCO dataset. We utilized a complete retrieval system (as depicted in Figure \ref{fig:retrieval_framework}), incorporating the FAISS \cite{johnson2019billion} search engine to enable a seamless online interactive retrieval. Additional qualitative examples are available in the Supplementary Materials.

\textbf{}
\vspace{-0.8cm}
\section{Conclusions}\label{sec:conclusions}

In this study, we present FOR, a finetuning framework designed for the task of object-centric open vocabulary image retrieval (OC-OVIR). FOR leverages SUM-CLIP, a specialized modified CLIP head tailored to capture multiple objects through additional learnable queries, combined with a multi-objective training approach. This training paradigm finetunes the model using available labels from the target dataset and pseudo-labels from a fixed CLIP-based architecture. It effectively mitigates the open vocabulary catastrophic forgetting problem and improves retrieval accuracy, particularly for novel categories unseen during training. Our approach yields enhanced retrieval rates while maintaining a compact image representation, thereby enabling efficient large-scale retrieval. Looking ahead, leveraging frameworks for learning compact open-vocabulary representations holds promise across various applications such as detection, segmentation, and image generation. We plan to further investigate these avenues and explore their integration with large language models in our future research.

{\small
\bibliographystyle{ieee_fullname}
\bibliography{egbib}
}
\newpage
\appendix
\section*{\LARGE{Supplementary Materials}}
\section{More Implementation Details}
Following \cite{LeviHLF23}, images were resized to a resolution of $448 \times 448$ for COCO and LVIS, and $768 \times 768$ for nuImages, with positional embeddings interpolated if needed. For fair comparison, we ensembled over the seven best CLIP prompts \cite{owlvit} in all CLIP-based models. Training was performed on a single Nvidia GPU (32GB RAM) and took an average of 16 hours. Since LVIS is a federated dataset wherein not all categories are annotated in each image, we augmented it with pseudo negatives. Specifically, we generated pseudo negatives by randomly sampling additional categories, weighted by their frequency in the dataset, to guarantee a minimum of 50 negatives and pseudo-negatives for supervised loss computation. Table \ref{tab:hyperparams} summarizes the hyper-parameters used in FOR. 
\begin{table}[h]
\captionsetup{font=footnotesize}
\centering
\footnotesize
\begin{tabular}{l|c|c|c}
\toprule
Hyperparameter & COCO & LVIS & nuImages \\
\midrule
Backbone & \multicolumn{3}{c}{ResNet50x64} \\
\# Decoder layers & \multicolumn{3}{c}{2} \\
\# Learnable queries & \multicolumn{3}{c}{50} \\
Learnable queries dimension & \multicolumn{3}{c}{4096} \\
Epochs & \multicolumn{3}{c}{25} \\
Learning rate & \multicolumn{3}{c}{1e-5} \\
Learning rate drop epoch & \multicolumn{3}{c}{15} \\
Learning rate drop factor & \multicolumn{3}{c}{0.1} \\
Batch size & \multicolumn{3}{c}{64} \\
Weight decay & \multicolumn{3}{c}{1e{-4}} \\
Optimizer & \multicolumn{3}{c}{Adam} \\
Adam $\beta_1$ & \multicolumn{3}{c}{0.9} \\
Adam $\beta_2$ & \multicolumn{3}{c}{0.999} \\
Dropout ratio & \multicolumn{3}{c}{0.1} \\
Pseudo labels conf. threshold & \multicolumn{3}{c}{5e-4} \\
\# KMeans clusters & \multicolumn{3}{c}{50} \\
$\gamma_{sup}$ & \multicolumn{3}{c}{1} \\
\midrule
$\gamma_{pse}$ & 1 & 1 & 10\\
Image resolution & $448$ & $448$ & $768$ \\
$w_\phi$ & 0.1 & 0 & 0 \\

\bottomrule 
\end{tabular}
\vspace{-0.2cm}
\caption{Hyperparameters used in FOR}
\label{tab:hyperparams}
\end{table}

\section{Comparison with Detection Frameworks}
Table \ref{tab:det} displays retrieval results on COCO-2017 validation set for various open vocabulary detectors suitable for the OVD-COCO benchmark\footnote{Detectors are fine-tuned on the COCO train set, except for OwlViT (fine-tuned on Objects365 \cite{Objects365} and Visual Genome \cite{VisualGenome}) and OwlV2 (fine-tuned on the LVIS base split) 
}, applied with their recommended settings and hyper-parameters. The `res' column indicates the image resolution, the `dual' column indicates the use of dual-encoder architecture, the `rep.' column indicates the total number of embeddings per image, and the `prop' column indicates the use of region proposals.

Notably, existing detection frameworks, including those deemed dual encoders, are impractical for retrieval tasks due to their inefficient embedding representation. Two stages detectors (BARON, CORA, CLIPSelf) rely on large number ($\sim$ 1000) of region proposals to enhance
novel category recall, while dense detectors (OwlViT, OwlV2) utilize huge visual embedding space. In contrast, FOR achieves superior performance on novel categories while utilizing over an order of magnitude fewer embeddings per image, and with no requirement for pixel level annotations.

\begin{table}[t]
\captionsetup{font=footnotesize}
\begin{center}
\resizebox{0.95\linewidth}{!}{%
\begin{tabular}{lllll|cc}
\hline
Method & res. & dual & prop & rep. & base & novel  \\
\hline
BARON \cite{BARON}, RN50 & 800 & \checkmark & \checkmark & 1000  & 94.00 & 76.69  \\
CORA \cite{wu2023cora}, RN50 & 800 & $\times$ & \checkmark & 1000 & 78.76 & 74.29  \\
CORA \cite{wu2023cora}, RN50x4 & 800  & $\times$ & \checkmark & 1000 & 86.27 & 78.63  \\
CLIPSelf \cite{wu2024clipself}, ViT-B/16 & 640 & \checkmark & \checkmark & 1000 & 95.14 & 87.31 \\
OwlViT \cite{owlvit}, ViT-B/16 & 768  & \checkmark & $\times$ & 2304 & 75.28 & 78.03  \\
OwlViT \cite{owlvit}, ViT-L/14& 768  & \checkmark & $\times$ & 3600 & 78.29 & 82.10  \\
OwlV2 \cite{OWLV2}, ViT-B/16 & 960 & \checkmark & $\times$ & 3600 & 89.98 & 87.15 \\
OwlV2 \cite{OWLV2}, ViT-L/14 & 1008 & \checkmark & $\times$ & 5184 & 90.92 & 87.27 \\
\hline\hline
FOR, RN50x64 & 448  & \checkmark & $\times$ & \textbf{25} & 88.97 & \textbf{89.17}  \\
FOR, RN50x64  & 448  & \checkmark & $\times$ & \textbf{50} & 88.14 & \textbf{89.17}  \\
\hline
\end{tabular}}
\end{center}
\vspace{-0.55cm}
\caption{Retrieval evaluation of detection frameworks on the COCO-2017 val set, reporting mAP@50.}
\label{tab:det}
\end{table}

\section{Multi-Label Classification}

The task of open-vocabulary multi-label classification, where a model identifies all relevant labels within an image, is related to but yet distinct from object-centric open-vocabulary image retrieval (OC-OVIR), where images are ranked based on an ad-hoc query. We note that the necessity for rapid image retrieval in response to ad-hoc queries precludes additional image processing for each new query, effectively limiting retrieval systems to a dual-encoder architecture - an imposition not present in classification tasks.

Despite the above considerations, Table \ref{tab:multilabel-rec} presents the results of \methodshort compared to fine-tuned open-vocabulary multi-label classification methods on the COCO-2014 dataset, using the first 65 lexically ordered categories as base and the remaining categories as novel. We compare against CONSE \cite{norouzi2013zero}, LabelEM \cite{akata2015label}, Fast0tag \cite{zhang2016fast}, LESA \cite{huynh2020shared}, and Generative-MLZSL \cite{gupta2023generative}, as early methods predating CLIP, as well as to CLIP \cite{radford2021learning}, CLIP-finetuned and DualCoOp \cite{sun2022dualcoop}, as concurrent or subsequent to CLIP (where the later being the current published SoTA). 
Models are assessed for Generalized Zero-Shot Learning (GZSL), which includes both base and novel categories, and Zero-Shot Learning (ZSL), which includes only novel categories. Evaluation metrics include the F1 score, measuring the accuracy of label ranking within each image, and mean Average Precision (mAP), which assesses the accuracy of image ranking for each label. 

Our framework outperforms prior work based on CLIP \cite{sun2022dualcoop} by 3-5 mAP points, indicating improved retrieval accuracy. The performance gap widens significantly when compared to zero-shot methods predating CLIP.
Although FOR is not specifically designed for multi-label classification, it achieves comparable ZSL F1 scores (indicative of classification accuracy) to DualCoOp and significantly outperforms CLIP. We attribute this success to FOR's ability to maintain CLIP’s vision-language alignment while optimizing for the target dataset, enabling high performance on novel categories unseen during training. 

\begin{table}[t]
\captionsetup{font=footnotesize}
\begin{center}
\resizebox{0.95\linewidth}{!}{%
\begin{tabular}{l|cc|cc}
\hline
& \multicolumn{2}{c|}{GZSL} & \multicolumn{2}{c}{ZSL} \\
Method & F1 (K=3) & mAP & F1 (K=3) & mAP \\
\hline
CONSE \cite{norouzi2013zero} & 19.6 & 7.7 & 18.4 & 13.2 \\
LabelEM \cite{akata2015label} &  6.7 & 4.0 & 10.3 & 9.6 \\
Fast0tag \cite{zhang2016fast} & 33.8 & 27.9 & 37.5 & 43.3 \\
LESA \cite{huynh2020shared} & 26.7 & 17.5 & 33.6 & 31.8 \\
Generative MLZSL \cite{gupta2023generative} & 44.1 & 33.2 & 43.5 & 52.2 \\
CLIP \cite{radford2021learning} & 50.2 & 68.1 & 48.3 & 75.8 \\
CLIP-FT & 43.9 & 66.7 & 47.4 & 78.6 \\
DualCoOp \cite{sun2022dualcoop} & \textbf{67.3} & 78.6 & \textbf{50.6} & 78.3 \\
FOR, ours & 56.4 & \textbf{81.3} & 50.5 & \textbf{83.0} \\
\end{tabular}}
\end{center}
\vspace{-0.55cm}
\caption{Open Vocabulary Multi-Label Classification on COCO2014}
\label{tab:multilabel-rec}
\vspace{-0.3cm}
\end{table}

\section{Pseudo-Labels Examples}

\begin{figure*}[h]
\centering
\includegraphics[width=0.85\linewidth]{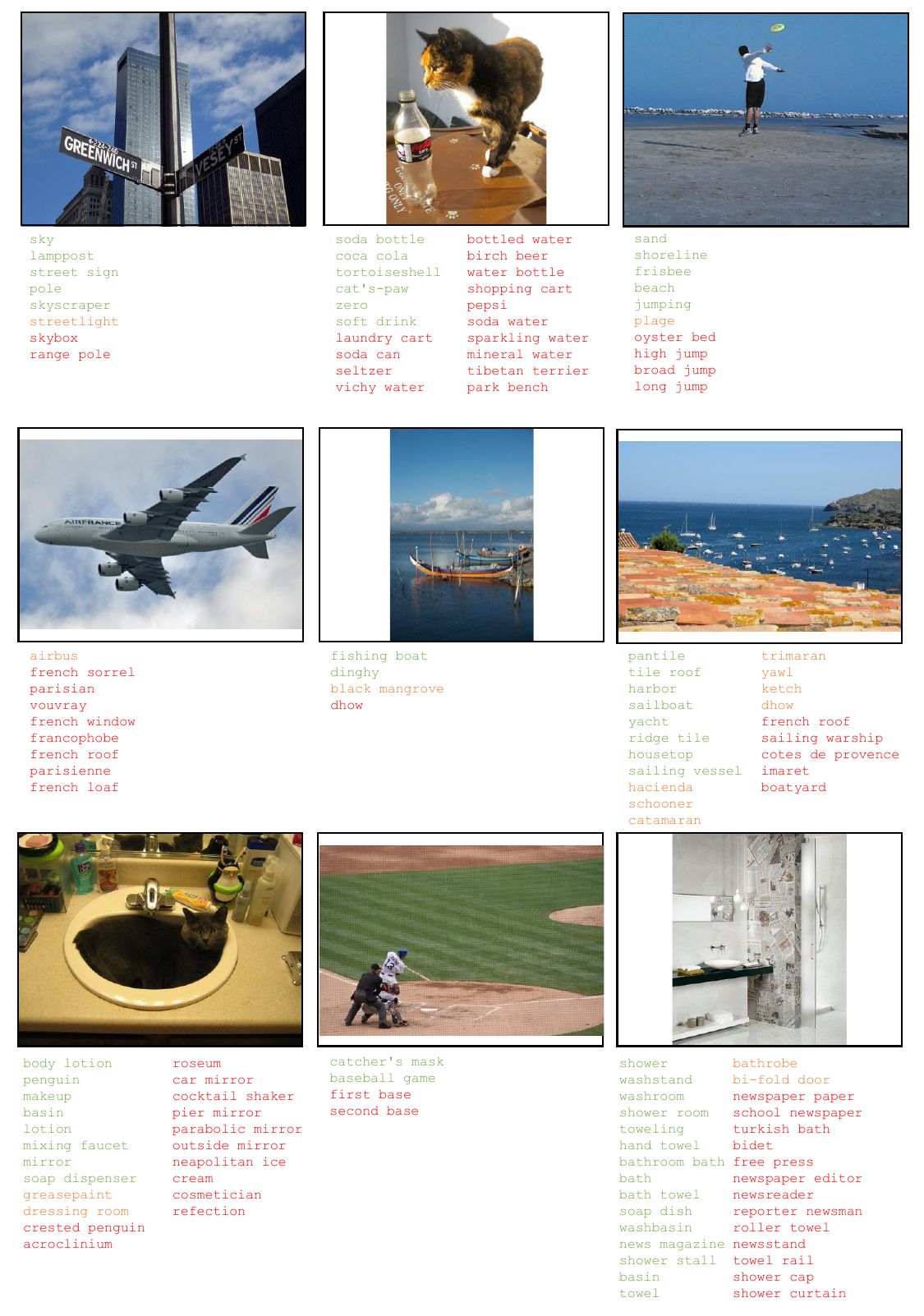}
\vspace{0.2cm}
\caption{\footnotesize\textbf{Pseudo-labels}: Example images from the COCO dataset with their generated pseudo labels. Pseudo labels are marked in \textcolor{green}{green}, \textcolor{orange}{orange}, and \textcolor{red}{red}, indicating labels that \textcolor{green}{exists}, \textcolor{orange}{might exists}, and \textcolor{red}{do not exists} in the image, respectively.}
\vspace{-0.5cm}
\label{fig:pseudo-labels}
\end{figure*}

Figure \ref{fig:pseudo-labels} presents visual examples from the COCO dataset, accompanied by their associated pseudo labels, which illustrate the advantages and limitations of the pseudo-labels assigned by \methodshort. Evidently, 
\methodshort manages to label most of the objects in the image, even for very small objects such as a frisbee (top row, right), sailing vessels (middle row, right), towels (bottom row, right), and cat's paws (top row, middle). Furthermore, pseudo-labels exist for more abstract concepts, such as a `baseball game' (bottom row, middle), as well as for out-of-distribution cases like a penguin in a bathroom (bottom row, left). 

The pseudo-labels include erroneous labels partially due to reliance on CLIP's vision-language embedding space and its limitations (Section \ref{supp:limitations}).  Additionally, errors can arise from common associations (e.g.,  an airplane image labeled 'AirFrance' yielding pseudo-labels like 'French loaf,' 'French roof,' and 'Parisian.') Finally, the model can mislabel objects by failing to detect context, such as identifying a wall with newspaper wallpaper as 'newspaper paper.'

\section{Overall System - Qualitative Examples}

\begin{figure*}[t]
\centering
\includegraphics[width=0.65\linewidth]{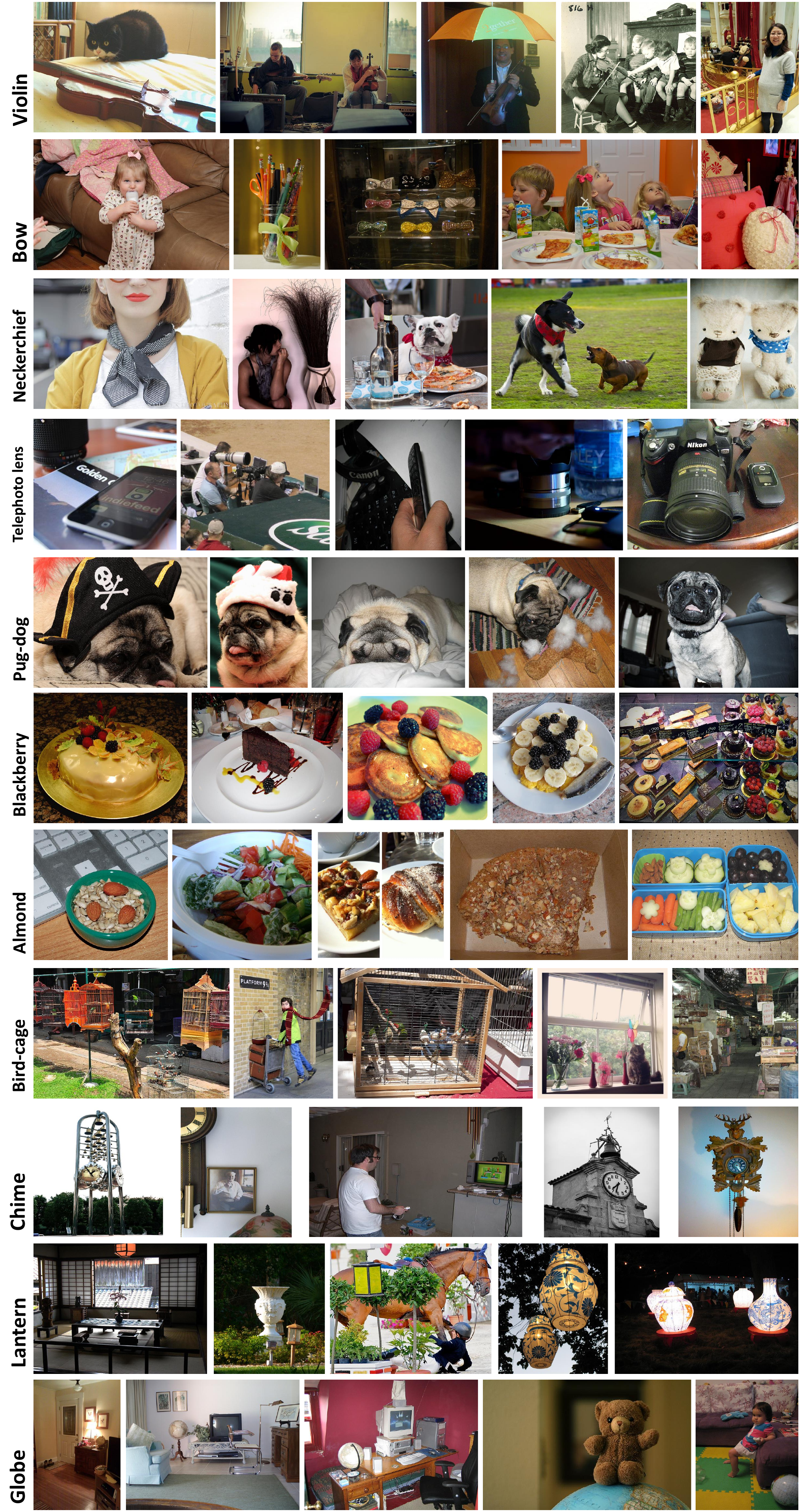}
\vspace{0.2cm}
\caption{\footnotesize\textbf{Qualitative examples of successful retrievals}: Top 5 images for various queries demonstrating \methodshort high retrieval rates, even for rare and challenging classes.}
\vspace{-0.5cm}
\label{fig:qual_ok}
\end{figure*}

\begin{figure*}[t]
\centering
\includegraphics[width=0.85\linewidth]{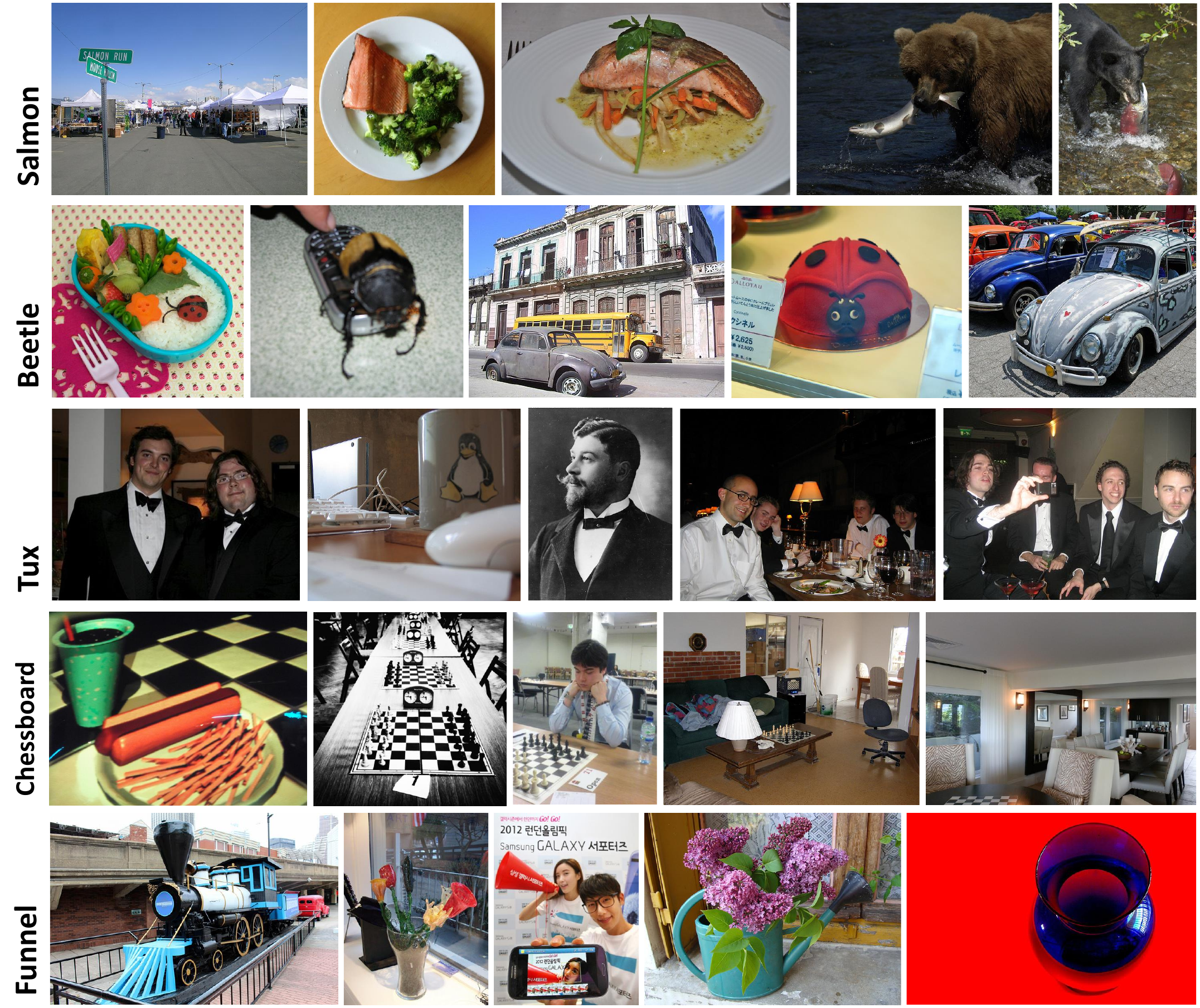}
\vspace{0.2cm}
\caption{\footnotesize\textbf{Qualitative examples illustrating error cases:} Top 5 retrieved images for various queries exemplifying the sensitivity of CLIP's embeddings space to text appearance ('salmon'), text ambiguity ('beetle'), and similarities in colors or shapes ('tux', 'chessboard,' and 'funnel'). Best viewed in color.}
\label{fig:qual_err}
\end{figure*}

\begin{figure*}[t]
\centering
\includegraphics[width=0.85\linewidth]{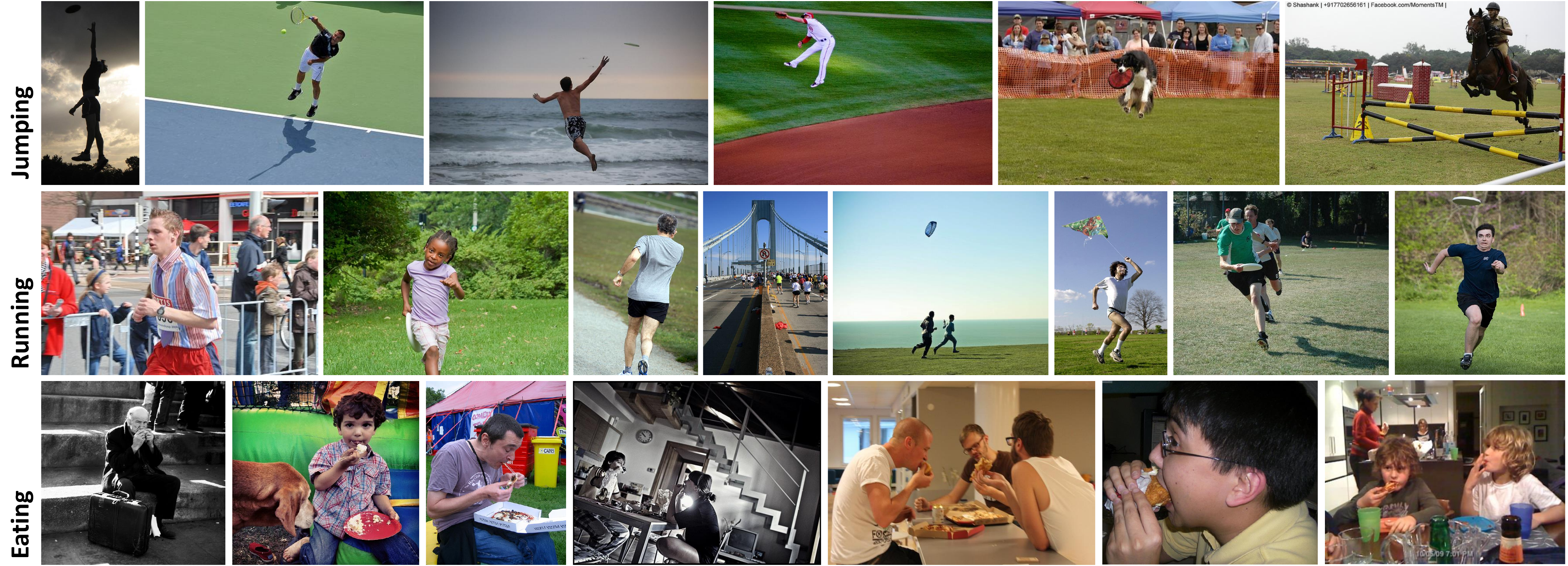}
\vspace{0.2cm}
\caption{\footnotesize\textbf{Qualitative examples, actions}: Evidence that \methodshort maintains CLIP's visual-language association.  
FOR is capable of retrieving images based on `action' queries, even after finetuning without specific `action' guidance. Best viewed with color.}
\vspace{1cm}
\label{fig:qual_action}
\end{figure*}

The overall retrieval system, illustrated in Figure \ref{fig:retrieval_framework} of the main article, facilitates iterative queries across extensive datasets. 
For demonstration, we integrated our fine-tuned visual encoder (utilizing a SUM-CLIP head generating 50 embeddings per image) and applied it to 120,000 images from the unlabeled segment of the COCO dataset. Qualitative results for diverse queries are  presented in Figures \ref{fig:qual_ok}, \ref{fig:qual_err} and \ref{fig:qual_action}. 
Figure \ref{fig:qual_ok} exhibits SUM-CLIP's high retrieval rates, presenting the top-5 retrieval results for text queries featuring rare classes such as `Violin', `Bow', `Globe' and others.
Figure \ref{fig:qual_err} illustrates error cases where the presence of text in the image (first row), text ambiguity (second row), similarities in colors (third row), patterns (fourth row) and shapes (fifth row), might lead to erroneous prioritization during the retrieval process.
Figure \ref{fig:qual_action} presents retrieval results for action queries (`jumping', `running' and `eating'). Despite our finetuning emphasis on objects, plausible qualitative results are achieved, indicating the retention of CLIP’s visual-language association in the fine-tuned model.

\section{Limitations}\label{supp:limitations}
Our approach, \methodshort, addresses the OC-OVIR task and as such has been tested primarily for the retrieval of images containing objects of interest. 
Initial experiments suggest that \methodshort has potential for non-object-centric retrieval tasks (see Figure \ref{fig:qual_action}), further testing could be conducted to assess its applicability in these contexts. Additionally, \methodshort builds upon the CLIP vision-language embedding space, inheriting its inaccuracies. Examples of interest are shown in Figure \ref{fig:qual_err}, where ranking is influenced by the presence of text (first row), or similarities in patterns or shapes (last rows). Lastly, CLIP is trained on noisy web data, which may inadvertently include private information, harmful text, and societal biases, all of which could potentially manifest during the application of our model.

\end{document}